\documentclass[a4paper]{article}
\usepackage{graphicx}
\usepackage{twocolceurws}
\usepackage{comment}

\title{Too Many Claims to Fact-Check: Prioritizing Political Claims Based on  Check-Worthiness}

\author{Yavuz Selim Kartal, Mucahid Kutlu, and Busra Guvenen \\
  Department of Computer Engineering \\
 TOBB University of Economics and Technology \\
  Ankara, Turkey \\
  \texttt{\{ykartal, m.kutlu, bguvenen\}@etu.edu.tr} }

\institution{}

\begin{document}

\maketitle

\begin{abstract}
The massive amount of misinformation spreading on the Internet on a daily basis has enormous negative impacts on societies. Therefore, we need automated systems helping fact-checkers in the combat against misinformation. In this paper, we propose a model prioritizing the claims based on their check-worthiness. We use BERT model with additional features including domain-specific controversial topics, word embeddings, and others. In our experiments, we show that our proposed model outperforms all state-of-the-art models in both test collections of CLEF Check That! Lab in 2018 and 2019. We also conduct a qualitative analysis to shed light detecting check-worthy claims. We suggest requesting rationales behind judgments are needed to understand subjective nature of the task and problematic labels.
\end{abstract}

\section{Introduction}
The World Economic Forum (WEF) has ranked massive digital misinformation as one of the top global risks in 2013\footnote{http://reports.weforum.org/global-risks-2013}. Unfortunately, the foresight of WEF seems right as we encountered many unpleasant incidents due to the misinformation spread on the Internet since 2013 such as the gunfight due to ``Pizzagate" fake news\footnote{www.nytimes.com/2016/12/05/business/media/comet-ping-pong-pizza-shooting-fake-news-consequences.html} and  increased mistrust towards vaccines\footnote{www.washingtonpost.com/news/wonk/wp/2014/10/13/the-inevitable-rise-of-ebola-conspiracy-theories}.

In order to combat against misinformation and its negative outcomes,  fact-checking websites (e.g., Snopes\footnote{https://www.snopes.com/})  detect the veracity of claims spread over the Internet and share their findings with their readers~\cite{cherubini2016rise}. However, fact-checking is an extremely time-consuming process, taking around one day for a single claim~\cite{Hassan2017ClaimBusterTF}. While these invaluable journalistic efforts help to reduce the spread of  misinformation, Vosoughi et al. \cite{vosoughi2018spread} report that  false news spread eight times faster than true news. 
Therefore, systems helping fact-checkers are urgently needed in the combat against misinformation.

As human fact-checkers are not able detect the veracity of all claims spread on the Internet, it is vital to spend their precious time in fact-checking the most important claims. Therefore, an automatic system  monitoring social media posts, news articles and statements of politicians, and detecting the \emph{check-worthy} claims is needed. A number of researchers  focused on this important problem (e.g., \cite{Hassan2017ClaimBusterTF,patwari2017tathya,jaradat2018claimrank}). Furthermore, Conference and Labs of Evaluation Forum (CLEF) Check That! Lab (CTL) has been organizing shared-tasks on detecting  check-worthy claims since 2018 ~\cite{nakov2018overview,atanasova2019overview,BarrnCedeo2020CheckThatAC}. In CTL tasks, a political debate or a transcribed speech is  separated by sentences and participants are asked to rank the sentences according to their priority to be fact-checked. In CTL'20 \cite{10.1007/978-3-030-58219-7_17}, tweets 
have also been used for this task. 

In this paper, we propose a ranking model that prioritizes claims based on their check-worthiness. We propose a BERT-based hybrid system 
in which we first fine tune a BERT~\cite{devlin2019bert} model 
for this task, and then use its prediction and other features we define in a logistic regression model to prioritize the claims.
The features we use include word-embeddings,  presence of comparative and superlative adjectives,  domain-specific controversial topics, and others. Our model achieves 0.255 and 0.176 mean average precision (MAP) scores in CTL'18 and CTL'19 datasets, respectively, outperforming all state-of-the-art models including participants of the corresponding shared-tasks, ClaimBuster~\cite{Hassan2017ClaimBusterTF}, BERT, XLNET \cite{yang2019xlnet}, and Lespagnol et al.\cite{lespagnol2019information}'s model. We share our code  for the reproducibility of our results\footnote{https://github.com/YSKartal/political-claims-checkworthiness}. 

\section{Related Work}\label{sec:rel}

As the US presidential election in 2016 is one of the main motivating reasons for fact-checking studies, prior work mostly 
used debates and other speeches of US politicians as their datasets (e.g., \cite{Hassan2017ClaimBusterTF,lespagnol2019information}). Therefore, the majority of studies focused on English. The Arabic datasets used in prior work (\cite{jaradat2018claimrank,nakov2018overview}) 
are just translations of English datasets.

ClaimBuster~\cite{Hassan2017ClaimBusterTF} is one of the  first studies about check-worthiness. ClaimBuster is a supervised model  using many features including part-of-speech (POS) tags, named entities, sentiment, and TF-IDF representations of claims. 
TATHYA~\cite{patwari2017tathya}  
uses topics,  POS tuples, entity history, and bag-of-words  as features. The topics are detected by LDA model trained on  transcripts of all
presidential debates from 1976 to 2016. 

Gencheva et al. \cite{gencheva2017context} propose a neural network model with a long list of sentence level and contextual features including sentiment, named entities, 
word embeddings, topics, contradictions, and others. 
Jaradat et al. \cite{jaradat2018claimrank} use roughly the same features with Gencheva et al., but extend the model for Arabic. In its followup work, Vasileva et al. \cite{vasileva2019takes} propose a multi-task learning model to detect whether a claim will be fact-checked by at least five (out of nine) pre-selected reputable fact-checking organizations.

CLEF has been organizing Check That! Labs (CTL) since 2018.
Seven teams participated in check-worthiness task of CTL'18. 
The participant teams used various learning models such as  recurrent neural network (RNN)~\cite{Hansen2018TheCT}, multilayer perceptron~\cite{Zuo2018AHR}, random forest (RF)~\cite{DBLP:conf/clef/AgezBLPM18}, k-nearest neighbor (kNN)~\cite{DBLP:conf/clef/GhanemMPR18a} and Support Vector Machine (SVM)~\cite{DBLP:conf/clef/YasserKE18} with different sets of features such as bag-of-words~\cite{Zuo2018AHR}, character n-gram~\cite{DBLP:conf/clef/GhanemMPR18a}, POS tags~\cite{Zuo2018AHR,Hansen2018TheCT,DBLP:conf/clef/YasserKE18}, verbal forms~\cite{Zuo2018AHR}, 
named entities~\cite{Zuo2018AHR,DBLP:conf/clef/YasserKE18}, 
syntactic dependencies~\cite{Zuo2018AHR,Hansen2018TheCT}, and word embeddings~\cite{Zuo2018AHR,Hansen2018TheCT,DBLP:conf/clef/YasserKE18}. On English dataset, Prise de Fer~\cite{Zuo2018AHR} team achieved the best MAP scores using almost every feature  mentioned before with SVM-Multilayer perceptron learning.

In 2019, 11 teams participated in  check-worthiness task of CTL'19. Participants used varying models such as LSTM, 
SVM, naive bayes, and logistic regression (LR) with many features including 
readability of sentences and  their context  
 ~\cite{atanasova2019overview}. Copenhagen team~\cite{DBLP:conf/clef/Hansen0SL19} achieved the best overall performance using syntactic dependency and word embeddings with weakly supervised LSTM model. 

Lespagnol et al. \cite{lespagnol2019information} investigated using various learning models such as SVM, LR, and Random Forests, 
with a long list of features including word-embeddings,  POS tags, syntactic dependency tags, entities, and ``information nutritional" features which represent  factuality, emotion, controversy, credibility, and technicality of statements. 
In our experiments we show that our model outperforms Lespagnol et al. on both test collections.

Our proposed an approach distinguishes from the existing studies as follows. 1) We propose a BERT-based hybrid model which uses fine-tuned BERT's output with many other features. 
2) As the topic might be a strong indicator for check-worthiness, many studies used various types of topics such as general topics 
\cite{DBLP:conf/clef/YasserKE18}, globally controversial topics \cite{lespagnol2019information}, and topics discussed in old US presindential debates~\cite{patwari2017tathya}. However, we believe that check-worthiness of a claim depends on local and present controversial topics. Thus, we use a list of  hand-crafted controversial topics related to US elections. 3) We also use two  different sets of features including a hand-crafted list of words and presence of comparative and superlative adjectives and adverbs.

\section{Proposed Approach}\label{sec:approach}
 We propose a supervised model with a number of features described below.
We investigate various learning models including LR, SVM, random forest, 
MART \cite{Friedman2001GreedyFA}, and LambdaMART \cite{10.1007/s10791-009-9112-1}. Now we explain the features we use.

\textbf{BERT:} We first fine tune BERT using respective training data. Next, we use its prediction value as one of our features. 

\textbf{Word Embeddings (WE):}
Words that are semantically and syntactically similar tends to be close in the embedding space, allowing us to capture similarities between claims. 
We represent a sentence as the average vector of its words excluding the out-of-vocabulary ones. 
Word embedding vectors are extracted from the pre-trained word2vec model~\cite{mikolov2013efficient} which has a feature vector size of 300.

 

\textbf{Controversial Topics (CT):}
Sentences about controversial topics might include check-worthy claims. Lespagnol et al. \cite{lespagnol2019information}  use a list of controversial issues compiled from Wikipedia article ``Wikipedia:List\_of\_controversial\_issues". However, the list they use covers many controversial issues which have very limited coverage in current US media such as ``Lebanon", ``Chernobyl", and ``Spanish Civil War" while the data we use are about recent US politics. We believe that controversy of a topic depends on the society. For instance, US politicians propose different policies for immigrants, yielding heated discussions among them and their supporters. On the other hand, US domestic politics are much less interested in refugee crisis in Mediterranean sea than European countries. Therefore, a claim about Mexican immigrants might be check-worthy for people living in US while they might find claims about refugees taking a dangerous path to reach Europe not-check-worthy. In contrast, people living in Europe might consider the latter case as check-worthy and the former one as not-check-worthy.
In addition, controversy of a topic might change over time. For instance, 
Cold War (which also exists in that Wikipedia list) might be one of the most discussed topics in US politics before  the collapse of the Soviet Union in 1991. However, nowadays  it is rarely covered by US media. Therefore, we propose using   controversial issues related to the data we use,  instead of any  controversial issue around the globe and in the history. 

Firstly, we identified 11 major topics in current US politics including immigration, gun policy,  racism, education, Islam, climate change, health policy, abortion, LGBT, terror, and wars in Afghanistan and Iraq. For each topic, we identified related words and calculate the  average of these words using their word embedding vectors.
For instance, for the immigration topic, we used words ``immigrants", ``illegal", ``borders", ``Mexican", ``Latino" and ``Hispanic". 

In this feature set of size 11, we calculate cosine similarity between sentences and each topic by using their vector presentation. We  use the average of word embeddings for sentences excluding stopwords with NLTK~\cite{Loper02nltk:the}.

\textbf{Comparative \& Superlative (CS):}
Politicians frequently use sentences comparing themselves with others because each candidate tries to convince the public that s/he is better than his/her opponent. Therefore, the comparisons in political speeches might  impact  people's voting decision and, thereby, it might be important to check their veracity.
Thus, in this feature, we  use the number of comparative and superlative adjectives and adverbs in sentences. 

\textbf{Handcrafted Word List (HW):} 
Particular words convey important information about check-worthiness because 1) it might be related to an important topic (e.g., ``unemployment"), 2) it represents a numerical value, increasing the factuality of the sentence (e.g., ``percent") and 3) its semantic represents a comparison between two cases (e.g., ``increase" and ``decrease"). Thus, we first identified 66 words analyzing training datasets of CTL'18 and CTL'19. In this feature, we check whether there is an overlap between lemmas of selected words and lemmas of words in the respective sentence. 

\textbf{Verbe Tense (VT):} We cannot detect the veracity of claims about future while we can only verify claims about the present or past. Thus, the verbe tense of sentences might be an effective indicator for check-worthiness of claims. This feature vector 
represents the existence or absence of each tense in the predicate of the claims. 

 \textbf{Part-of-speech (POS) Tags:} 
If a sentence does not contain any informative words, then it is less likely to be check-worthy. To represent the information load of a claim, we use the  number of nouns, verbs, adverbs and adjectives, separately. 

\section{Experiments}\label{sec:experiments}

\subsection{Experimental Setup}

\textbf{Implementation:}  We use  ktrain library\footnote{https://pypi.org/project/ktrain/}  to fine-tune BERT model 
 with 1 cycle learning rate policy and maximum learning rate of 2e-5~\cite{Smith2018ADA}. We use SpaCy\footnote{https://spacy.io/} for all syntactic and semantic analyses. 
 We use Scikit toolkit\footnote{https://scikit-learn.org} for the implementations of SVM, Random Forest (RF), and LR. The parameter settings of the learning algorithms are as follows. We use default parameters for SVM. We set the number of trees to 50 and the maximum depth to 5 for RF. We use multinomial and lbfgs settings for LR. 
 For MART and LambdaMART models, we use RankLib\footnote{https://sourceforge.net/p/lemur/wiki/RankLib/} library, and set the number of trees and leaves to 50 and 2, respectively.

\noindent
\textbf{Data:} We evaluate the performance of our system with two datasets used in CTL'18 and CTL'19. The details about them are given in \textbf{Table~\ref{table_dataset}}. CTL'18 consists of transcripts of debates and speeches while CTL'19 contains also press conferences and posts.  

\begin{table}[h!]
  \begin{center}
    \caption{Details about CTL'18 and CTL'19 datasets. 
    }
    \label{table_dataset}
    \begin{tabular}{|c|l|c|c| c||c|c|c|} \hline
        
       & \textbf{} &  \textbf{CTL'18}  &  \textbf{CTL'19}  \\ \hline
      
     & \textbf{\# Docs} & 3 & 19 \\
      \textbf{Train} & \textbf{\# Sentence} &  {4,064} &  {16,421} \\
      & \textbf{\# CW Claims} & {90 (2,2\%)} & {433 (2,6\%)} \\ \hline
      
        & \textbf{\# Docs} & 7 & 7  \\ 
        
        \textbf{Test}& \textbf{\# Sentence} &  {4,882} & {7,079}  \\ 
        & \textbf{\# CW Claims} & {192 (3,9\%)}   &{110 (1,6\%)}  \\  \hline
        
    \end{tabular}
  \end{center}
\end{table}

\noindent
\textbf{Baselines:} We compare our model against the following models.
\begin{itemize}
\item \textit{Lespagnol et al. \cite{lespagnol2019information}}:   Lespagnol et al. report the best  results on CTL'18 so far. Therefore, we use it as one of our baselines. In order to get its results for CTL'19, we contacted with the authors to get their own code. The authors provide us the values of ``information nutrition" features and instructions about how to generate WE embeddings. We implemented their method using the values they shared and following their instructions\footnote{It is noteworthy that we obtain 0.2115 MAP score on CTL'18 with our implementation of their method while they report 0.23 MAP score in their paper. We are not aware of any bug in our code but the performance difference might be because of different versions of the same library. Nevertheless, the results we present for their method on CTL'19 should be taken with a grain of salt.}.  
\item \textit{ClaimBuster}: We use the popular pretrained ClaimBuster API\footnote{https://idir.uta.edu/claimbuster/} \cite{Hassan2017ClaimBusterTF} which is trained on a dataset  covering different debates that do not exist on CTL'18 and CTL'19. 

\item \textit{BERT}: As it is reported that BERT based models outperform  state-of-the-art models in various NLP tasks, we compare our model against using only BERT. We fine tune BERT model using the respective training dataset and predict the check-worthiness of claims using the fine-tuned model.

\item \textit{XLNET}: It is reported that XLNet outperfroms BERT in various NLP tasks \cite{yang2019xlnet}. Thus, we use XL-NET for this task by fine-tuning with the  respective training dataset.  

\item \textit{Best of CTL'18 and CTL'19}: For each dataset, we also report the performance of best systems participated in the shared-tasks, i.e., Prise de Fer team~\cite{Zuo2018AHR} and Copenhagen team~\cite{DBLP:conf/clef/Hansen0SL19}  for CTL'18 and CTL'19, respectively.
\end{itemize}

\noindent
\textbf{Training \& Testing:} We use the same setup with CTL'18 and CTL'19 to maintain a fair comparison with the baselines. 
We follow the evaluation method used on CTL'18 and CTL'19: We calculate average precision (AP), 
R-precision (RP), precision@5 (P@5) and precision@10 (P@10) for each file (i.e., debate, speech) and then report the average performance.
 

\subsection{Experimental Results}

In this section, we present experimental results on test data using different sets of features and varying learning algorithms. 

\noindent
\textbf{Comparison of Learning Algorithms.} In our first set of experiments, we evaluate logistic regression (LR), SVM, random forest (RF), MART and LambdaMART models using all features defined in Section \ref{sec:approach}.  \textbf{Table~\ref{tab:test:models}} shows MAP scores of each model. Interestingly, LR outperforms all other models. In a similar experiment Lespagnol et al.\cite{lespagnol2019information} conducted, they also report that LR yields higher results than other models they used. Nevertheless, 
we use LR in our following experiments.   

\begin{table}[h!]
  \begin{center}
    \caption{\textbf{MAP Score for Varying Models Using All Features}}
    \label{tab:test:models}
    \begin{tabular}{|c|c|c|}
    \hline
      \textbf{Learning Model} & \textbf{ CTL'18 } & \textbf{ CTL'19  } \\ 
      \hline
     \textbf{LR} & \textbf{.2303}  & 	\textbf{.1775}   \\ \hline
     \textbf{RF} & .1468  & 	.1542   \\ \hline
     \textbf{SVM} & .1716	 & .1346	   \\ \hline
     \textbf{MART} & .1764	 & .1732	   \\ \hline
     \textbf{Lambda MART} & .0671	 & .0564	   \\ \hline
    \end{tabular}
  \end{center}
\end{table}

\noindent
\textbf{Feature Ablation.} 
In order to analyze the effectiveness of features we use, we apply two techniques: 1) \emph{Leave-one-out methodology} in which we exclude one type of feature group and calculate the model's performance without it, and 2) \emph{Use-only-one methodology} in which only a single feature group is used for prediction.
The results are shown in \textbf{Table~\ref{tab:test:feature}}.

\begin{table*}[htb]
  \begin{center}
    \caption{\textbf{MAP Scores for Varying Feature Sets}}
    \label{tab:test:feature}
    \begin{tabular}{|l|c|c||l|c|c|}
    \hline
     \multicolumn{3}{|c||}{ \textbf{Leave-One-Out} } & \multicolumn{3}{c|}{ \textbf{Use-Only-One} } \\ \hline
      \textbf{Features}  & \textbf{ \textbf{CTL18}} & \textbf{ \textbf{CTL19}} &  \textbf{Features}  & \textbf{ \textbf{CTL18}} & \textbf{ \textbf{CTL19}} \\ \hline
       \textbf{All}  & {.2303}  & .1775  &  &  &  \\ \hline
       \textbf{All-CS}  & {.2239}    & .1765 &    \textbf{CS} & .751 & .604	    \\ \hline
       \textbf{All-BERT}  & {.2211}  & .1580 &  \textbf{BERT} & {.1850}	 & \textbf{.1701}  \\ \hline
      \textbf{All-VT}  & \textbf{.2547}  & .1761 & \textbf{VT} & {.1007}  & 	{.598}  \\  \hline
      \textbf{All-HW}  & {.2126} & .1727 & \textbf{HW} & .1530  & 	.1043  \\ \hline
      \textbf{All-WE}  & {.1756}   & \textbf{.1786} & \textbf{WE} & \textbf{.2068}  & 	{.1356}  \\ \hline
      \textbf{All-CT}  & {.2170}   & .1739 & \textbf{CT} & .1363  & 	.1046 \\ \hline
      \textbf{All-POS}  & {.2283}    & .1767 & \textbf{POS} & .1048	 & .631   \\ \hline
    \end{tabular}
  \end{center}
\end{table*}

From the results in Table  \ref{tab:test:feature}, we see that features have different effects on each dataset. BERT is the most effective feature on CTL'19. However, in contrast to our expectations, WE seems more effective feature than BERT on CTL'18. 
On CTL'18,  
the performance decreases by nearly 25\%  when WE is excluded. In addition, we achieve the highest MAP score when we use only WE. On CTL'19, we achieve 0.1356 MAP score using only WE, showing that it is more effective than other features except BERT. However, the performance of our model increases when we exclude WE (0.1775 vs. 0.1786 in Table~\ref{tab:test:feature}), suggesting that the information it contributes is covered by other features on CTL'19.

\begin{table*}[!htb]
  \begin{center}
    \caption{\textbf{Comparison with Competing Models. 
    * sign indicates the results obtained from our implementation of the respective competing model.}}
    \label{tab:test}
    \begin{tabular}{|l|c|c|c|c||c|c|c|c|}
    \hline
     & \multicolumn{4}{c||}{\textbf{CTL'18}} & \multicolumn{4}{c|}{\textbf{CTL'19}} \\ \cline{2-9}
      \textbf{Model}  & \textbf{ MAP} & \textbf{ RP }& \textbf{P@5 } & \textbf{P@10} & \textbf{ MAP} & \textbf{ RP }& \textbf{P@5 } & \textbf{P@10}\\ \hline
       \textbf{BERT}  & {.1850}  & 	.2218 &  .3142 &  {.2857}   & .1701  & 	.1945 &  \textbf{.2571} & \textbf{.2429}  \\ \hline
       
        \textbf{XLNET}  & {.1974}  & 	.2393 &  .2857 &  {.2571}   & .0932  & 	.0770 &  {.1429} & {.1143}  \\ \hline
       \textbf{Lespagnol et al. \cite{lespagnol2019information}  }  & {.230}  & 	.254  & .314 & .2857*  & .1292*  & 	.1347* & {.1714*} & .2000*  \\ \hline
       \textbf{Prise de Fer Team} & {.1332}  & 	.1352  & .2000 & {.1429} &-&-&-&-\\ \hline
       \textbf{Copenhagen Team}  & - & 	-  & - & -   & .1660  & 	\textbf{.4176} & \textbf{.2571} & .2286  \\ \hline
      \textbf{ClaimBuster}  & {.2003}  & 	.2162 & .2571 & {.2429}   & .1329  & 	{.1555}  & {.1714} & .2000  \\ \hline
      \hline
      \textbf{Our Model}  & \textbf{.2547}  & 	\textbf{.2579} &  \textbf{.4000} & \textbf{.3429}   & \textbf{.1761}  & 	.2028 &  \textbf{.2571} & .2143  \\ \hline
    \end{tabular}
  \end{center}
\end{table*}

Excluding hand-crafted word list (HW) features causes performance decrease in both test collections. In addition, using only HW features outperforms all participants of CTL'18 (0.153 vs 0.1332 in Table~\ref{tab:test:feature}). These promising results suggest that expanding this list might lead further performance increases.

Our results also suggest that Controversial Topics (CT) are effective features. Excluding them decreases the performance of the model in both collections while using only CT features yield high scores, slightly outperforming the best performing system on CTL'18 (0.1363 vs. 0.1332 in Table~\ref{tab:test:feature}).

Excluding CS and POS features also slightly decrease the performance of the model in both test collections. Regarding time tense features, our results are mix. Excluding time tense feature causes a slight performance decrease on CTL'19, but yields higher performance score on  CTL'18. 


\noindent
\textbf{Comparison Against Baselines.}
We pick the model that includes all features except VT 
as our primary model because it achieves the highest MAP score on average. We compare our primary model with the baselines. The results are presented in \textbf{Table~\ref{tab:test}}. 

Our proposed model outperforms all other models based on all evaluation metrics on CTL'18. On CTL'19, our proposed model achieves the highest MAP score, which is the official metric used in CTL. BERT model outperforms other models based on P@10 on CTL'19. Regarding P@5 metric, our model, BERT and Copenhagen Team achieve the same highest scores with 0.2571. Regarding RP, Copenhagen Team achieves the highest score. 
Overall, our model outperforms all other models based on the official evalution metric of CTL while BERT and Copenhagen Team~\cite{Hansen2018TheCT} also achieve comparable performance on CTL'19.

\begin{table*}[htb]
\small
  \begin{center}
    \caption{Highest ranked non check-worthy statements from each test document by our primary model}
    \label{tab:qualitative}
    \begin{tabular}{|c|c|p{2.55cm}|c|p{8cm}|}
    \hline
    \textbf{Row}  &  \textbf{Rank} & \textbf{File  Name}   & \textbf{ Speaker } & \textbf{ Statement }  \\ 
      \hline
      1& 4 & task1-en-file1 & CLINTON & The plan he has will cost us jobs and possibly lead to another Great Recession.    \\ 
      \hline
      2& 1 & task1-en-file2 &CLINTON & Then he doubled down on that in the New York Daily News interview, when asked whether he would support the Sandy Hook parents suing to try to do something to rein in the advertising of the AR-15, which is advertised to young people as being a combat weapon, killing on the battlefield.     \\ 
      \hline
      3& 1 & task1-en-file3 &TRUMP & Jobs, jobs, jobs.    \\ 
      \hline
      4& 2 & task1-en-file4 &TRUMP & Before that, Democrat President John F. Kennedy championed tax cuts that surged the economy and massively reduced unemployment.     \\ 
      \hline
      5& 3 & task1-en-file5 &TRUMP & The world's largest company, Apple, announced plans to bring \$245 billion in overseas profits home to America.     \\ 
      \hline
      6& 1 & task1-en-file6 &TRUMP & America has lost nearly-one third of its manufacturing jobs since 1997, following the enactment of disastrous trade deals supported by Bill and Hillary Clinton.     \\ 
      \hline
      7& 1 & task1-en-file7 &TRUMP & Our trade deficit in goods with the world last year was nearly \$800 billion dollars.     \\ 
      \hline
      8&1 & 20151219\_3\_dem &O'MALLEY & We increased education funding by 37 percent.     \\ 
      \hline
      9&1 & 20160129\_7\_gop &KASICH & We're up 400,000 jobs.     \\ 
      \hline
      10&1 & 20160311\_12 \_gop &TAPPER & Critics say these deals are great for corporate America's bottom line, but have cost the U.S. at least 1 million jobs.     \\ 
      \hline
      11&3 & 20180131\_state \_union &TRUMP & Unemployment claims have hit a 45-year low.      \\ 
      \hline
      12&1 & 20181015\_60\_min &TRUMP & --if you think about it, so far, I put 25\% tariffs on steel dumping, and aluminum dumping 10\%.     \\ 
      \hline
      13&3 & 20190205\_trump \_state &TRUMP & Unemployment for Americans with disabilities has also reached an all-time low.     \\ 
      \hline
      14&1 & 20190215\_trump \_emergency &TRUMP & They have the largest number of murders that they've ever had in their history - almost 40,000 murders.     \\ 
      \hline
        \end{tabular}
  \end{center}
\end{table*}

\section{Qualitative Analysis}\label{sec:qual}

In this section, we present our qualitative analysis for the output of our primary model. 
 For each input file, we rank the claims based on their check-worthiness and then detect not-check-worthy claim with the highest rank. \textbf{Table~\ref{tab:qualitative}} shows these not-check-worthy statements for each file with our system's ranking and speaker of the statement.

The statement in Row 1 is a claim about the future. Our model with verb tense could rank this statement at lower ranks but our primary model does not use verb tense features because it yields lower performance on average. In Row 2, the statement is very complex with many relative clauses, in perhaps decreasing the performance of  BERT model and WE features in representing the statement. In Row 3, our model makes an obvious mistake and ranks a statement which  does not have even any predicate, at very high ranks. Perhaps our model falls short because the word ``jobs" indicates that the statement is about  unemployment, which is one of the controversial topics we defined.

As reported by Vasileva et al. \cite{vasileva2019takes} fact-checking organizations investigate different claims with very minimal overlaps between selected claims. We observe this subjective nature of annotations in Rows 4-14 because all statements are actually factual claims and some of them might also be  considered as check-worthy. For instance,  statements in Row 8, 11 and 13 are clearly said to change people's voting decision.  In addition, almost all statements are  about economics which is an important factor on people's votes. Therefore,  checking their veracity might be also important not to misinform public. Nevertheless, these examples show the 
the subjective nature of check-worthiness annotations. 


In addition to subjective judgments, we also noticed inconsistencies within the annotations. For instance, the statement in Row 9 (``We are up 400,000 jobs") also exists  in ``20160311\_12\_gop" file but annotated as ``check-worthy". In addition, there exists semantically very similar statements with different labels. For instance, Donald Trump's statement ``I did not support the war in Iraq" in $1079^{th}$ line of 20160926\_1pres file is labeled as ``not-check-worthy" while his statement in $1086^{th}$ line of the same file ``I was against the war in Iraq" is labeled as ``check-worthy". Both statements have similar meanings and exists in the same context (i.e., their position in file are very close). Therefore, both might have the same labels. As a counter argument, ``being against" suggests an action while ``not supporting" does not require any action to be taken. Thus, different annotations for similar statements might also be again due to the subjective nature of check-worthiness judgments.


Furthermore, there are also annotations that we strongly disagree with the label. For instance, in 20170315\_nashville file (training data on CTL'19), Donald Trump's statement ``We're going to put our auto industry back to work" is labeled as check-worthy. However,  the statement is about future and cannot be verified. 

Overall, our qualitative analysis suggests that annotating check-worthiness of claims is a subjective task and the annotations might be noisy. Kutlu et al. \cite{kutlu2018crowd} show that using text excerpts within documents as rationales help understanding disagreements in relevance judging. Similarly, we might request rationales behind check-worthiness annotations to 
understand if the label is due to a human judging error or the subjective nature of the annotation task. Furthermore, rationales behind these annotations might help us develop
 effective solutions for this challenging problem.

\section{Conclusion} \label{sec:conc}

In this paper, we presented a supervised method which prioritize claims based on check-worthiness.  We use logistic regression classifier with features including state-of-the-art language model BERT, domain-specific controversial topics, pretrained word embeddings, handcrafted word list, POS tags and comparative-superlative clauses. In our experiments on CTL'18 and CTL'19, we show that our proposed model outperforms all state-of-the-art models in both collections. We show that BERT's performance can be increased by using additional features for this task. In our feature ablation study, BERT model and word embeddings appear to be the most effective features while handcrafted word list and domain-specific controversial topics also seem effective. Based on our qualitative analysis, we believe requesting rationales for the check-worthiness annotations would further help in developing effective systems.


In the future, we plan to work on weak supervision techniques  to extend the training dataset. With the increased  data, we will be able explore using deep learning techniques for this task. 
In addition, we plan to extend our study to detect check-worthy claims in social media platforms because it is the channel where most of the people affected by misinformation. 
Moreover, working on different languages and building a multilingual model is an important research direction in the  combat against misinformation.

\bibliographystyle{abbrv}

\end{document}